%% file: 0-main.tex
\pdfoutput=1
\documentclass[conference]{IEEEtran}
\IEEEoverridecommandlockouts

\input{0-preamble.tex}

\begin{document}

\linespread{0.96}

\title{Composing Meta-Policies for Autonomous Driving \\ Using Hierarchical Deep Reinforcement Learning\vspace{-5pt}}

\author{%
Richard Liaw$^{2}$,
Sanjay Krishnan$^{2}$,
Animesh Garg$^{2,3}$,
Daniel Crankshaw$^{2}$,
Joseph E. Gonzalez$^{2}$,
Ken Goldberg$^{1,2}$
\thanks{
$^{1}\,$IEOR, $^{2}\,$EECS, UC Berkeley, CA USA; 
$^{3}\,$CS, Stanford University, Stanford CA US; 
The AUTOLAB at UC Berkeley (automation.berkeley.edu);
\newline
\texttt{\{rliaw, sanjaykrishnan, animesh.garg, jegonzal, goldberg\}@berkeley.edu, crankshaw@eecs.berkeley.edu}}%
}

\maketitle

\begin{abstract}
Rather than learning new control policies for each new task, 
it is possible, when tasks share some structure, to compose a "meta-policy" from previously learned policies.
This paper reports results from experiments using Deep Reinforcement Learning on a continuous-state, discrete-action autonomous driving simulator. We explore how Deep Neural Networks can represent meta-policies that switch among a set of previously learned policies, specifically in settings where the dynamics of a new scenario are composed of a mixture of previously learned dynamics and where the state observation is possibly corrupted by sensing noise. We also report the results of experiments varying dynamics mixes, distractor policies, magnitudes/distributions of sensing noise, and obstacles.
In a fully observed experiment, the meta-policy learning algorithm achieves 2.6x the reward achieved by the next best policy composition technique  with 80\% less exploration.
In a partially observed experiment, the meta-policy learning algorithm converges after 50 iterations while a direct application of RL fails to converge even after 200 iterations.
\end{abstract}


\input{1-introduction.tex}
\input{2-relatedwork.tex}
\input{3-fo-case.tex}
\input{4-po-case.tex}
\input{5-exp-setup.tex}
\input{6-exp-results.tex}
\input{7-conclusion.tex}
\input{6a-discussion.tex}

{\footnotesize
\section*{Acknowledgment}
Any opinions, findings, and conclusions or recommendations expressed in this material are those of the author(s) and do not necessarily reflect the views of the Sponsors.
This research was performed in part in the AUTOLAB at UC Berkeley in
affiliation with the AMP Lab, BDD, BAIR, the CITRIS "People and
Robots" (CPAR) Initiative: http://robotics.citris-uc.org
Multilateral Manipulation by Human-Robot Collaborative Systems and by Google, Cisco, Siemens, Cloudminds.
}

\vspace{-5pt}
\bibliographystyle{IEEEtranS}
\input{0-main.bbl}

\end{document}

%% file: 0-preamble.tex
\usepackage{graphics}
\usepackage[pdftex]{graphicx}
\usepackage[font={small}]{caption}
\DeclareGraphicsExtensions{.pdf,.png,.jpg}

\usepackage[font={small}]{caption}
\usepackage{subcaption}
\usepackage[rightcaption]{sidecap}
\usepackage{pbox}

\usepackage{mathtools}
\usepackage{amsmath, amssymb, amscd}
\usepackage{ wasysym } 
\usepackage{amsfonts}

\usepackage[ruled,vlined,linesnumbered]{algorithm2e} 
\usepackage{algorithmicx, algpseudocode} 

\DeclareMathAlphabet{\mathcal}{OMS}{lmsy}{m}{n}
\DeclareSymbolFont{largesymbols}{OMX}{cmex}{m}{n}
\usepackage{textcomp} 

\usepackage{array} 
\usepackage{tabularx}
\usepackage{multirow}
\usepackage{multicol}

\usepackage[T1]{fontenc} 
\usepackage[utf8]{inputenc}
\usepackage[english]{babel} 
\usepackage{units}
\usepackage{bm}
\usepackage{times} 
\usepackage{xspace}
\usepackage{flushend}
\usepackage{csquotes}
\usepackage{makeidx}

\let\labelindent\relax 
\usepackage[inline,shortlabels]{enumitem}

\usepackage{soul} 
\usepackage{subfiles} 

\usepackage[protrusion=true,expansion=true]{microtype}
\usepackage[yyyymmdd]{datetime}

\date{\protect\formatdate{1}{1}{2001}}
\makeatletter
\let\NAT@parse\undefined
\makeatother
\usepackage[numbers]{natbib}

\usepackage{url}
\makeatletter
\g@addto@macro{\UrlBreaks}{\UrlOrds}
\makeatother
\usepackage{color}
\usepackage[usenames,dvipsnames,table]{xcolor}

\usepackage{soul} 

\newcommand{\todo}[1]{\textcolor{red}{[#1?]}}

\newcommand{\ignore}[1]{}

\newcommand{\figref}[1]{Figure~\ref{fig:#1}}

\newcommand{\seclabel}[1]{\label{sec:#1}}
\newcommand{\secref}[1]{Section~\ref{sec:#1}}

\newtheorem{problem}{Problem}


%% file: 1-introduction.tex
\section{Introduction}

Consider the problem of designing a cruise control policy for a car.
The manufacturer has an analytic model for a control policy $\pi_{0}$ that works on new cars.
However, the manufacturer notices that as older cars wear, the controller is no longer effective.
The manufacturer then collects data for a prototypical 5-year-old car at the end of the warranty cycle; identifies the system from data, and arrives at a policy $\pi_{5}$.
Given a car of the same make/model but with an unknown age, how can the manufacturer leverage $\pi_{0}$ and $\pi_{5}$ to control it? 

One approach would be to ignore $\pi_{0}$ and $\pi_{5}$ entirely, and apply a model-free control approach, such as Reinforcement Learning, to directly learn a new policy.
While this may in principle converge to the desired result, RL could spend an extensive amount of time exploring before learning.
However, a 6-month old car's cruise control policy may not be very different from $\pi_{0}$.
Instead of learning a policy from scratch, it may be better to selectively apply $\pi_{0}$ or $\pi_{5}$ in certain circumstances, e.g., at higher speeds the car behaves more like an old car than a new one. 
This issue can be addressed by a meta-policy since it will adaptively learn when to use $\pi_{0}$ or $\pi_{5}$.
Instead of applying RL over the original action space, we can apply RL to select between the two choices of applying either $\pi_{0}$ or $\pi_{5}$.
The reduction of action space to $k$ (in general) discrete choices greatly reduces the complexity of many problems and allows for quicker convergence.

Composing previously learned policies to address a novel task is an important problem in robot learning. One may be able to collect data for many specific tasks, e.g., driving in different conditions or over different terrain types.
From this data, we train a set of \emph{policies} to solve the specific tasks, and the \textbf{policy composition problem} is given a new task, learn a policy that is composed from the set.  
This has been studied as a variant of Hierarchical Reinforcement Learning (HRL)~\cite{kaelbling1993hierarchical, barto2003recent, comanici2010optimal, konidaris2009efficient, taylor2009transfer, dietterich2000hierarchical}, where a Reinforcement Learning algorithm learns a state-dependent meta-policy that switches between a set of closed-loop controllers called options.
We take inspiration from this prior work and evaluate how these approaches can be integrated with new results from Deep Reinforcement Learning~\cite{mnih2013playing}.

\begin{figure}[t]
\centering
 \includegraphics[width=\linewidth]{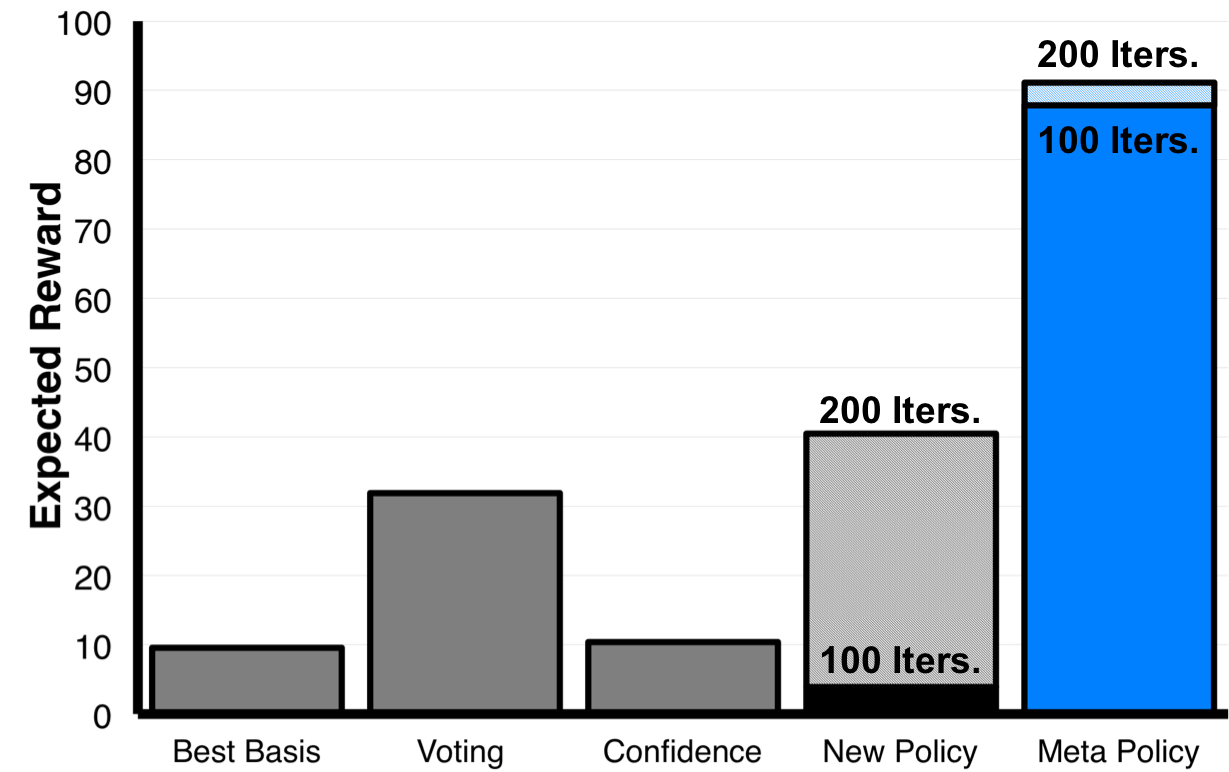}
 \caption{We evaluate the use of meta policies on a car simulator from~\cite{geramifard}. Two policies are trained on known dynamical regimes and the goal is to control the car given an unknown composition of these regimes. The best of the two policies achieves a reward of 9.55, a voting ensembling technique achieves 31.92, and a confidence ensembling technique achieves 10.32. Training a new policy with RL achieves a reward of 89.16 after 500 iterations of exploration, while meta policy learning achieves a comparable reward 87.90 after only 100 iterations.}
 \label{fig:baselines}
 \vspace{-15pt}
\end{figure}

To formalize the problem, suppose we are given a set of policies $\{\pi_1,...,\pi_k\}$ which we call \emph{basis policies}, and all we know about each policy is that it is a function from a continuous state space $S$ to discrete action set $A$.
We assume that we can query each of the policies for actions at any state $s \in S$.
Given a new task specified by a transition function $T(s,a)$ and reward function $R(s,a)$, the objective is to find an optimal meta policy $\pi^{meta}: S \mapsto \{1, \ldots, k \}$ that at each state selects one of the $k$ policies to apply.
Even though there are a discrete number of basis policies, the selection function can be quite complex and difficult to represent exactly.
One solution is to parametrize this function $\pi^{meta}_\theta$ with a highly expressive function class such as a Neural Network.
Recent research in Deep RL has proposed many techniques to learn such parametrized policies~\cite{schulman2015trust}. 

One challenge in RL is the problem of partial observation in the form of dropped and/or noisy sensor measurements.
In the cruise control example, consider the case when older cars have noisier velocity sensors.
While RL is robust to process noise, sensing noise (where the robot cannot directly observe the state of the world) remains to be a challenge.
This leads to policies that depend on history and not just the current state.
Fortunately, intriguing new results in Recurrent Neural Networks (RNNs) can address this problem.
RNNs are neural networks that have cycles, allowing it to capture temporal events that depend on the past.
RNNs are particularly useful in the context of partial observation since the network can learn which previous states to store in memory.
We 
use an RNN architecture based on Gated Recurrent Units (GRU) to represent the meta policies and the basis policies~\cite{chung2014empirical}.

We experimentally evaluate these techniques in the context of a continuous-state, discrete-action driving simulator~\cite{geramifard}.
We explore how DL-based meta policies can address driving tasks where the dynamics of the new task are unknown but composed of a mixture of previously seen dynamics, when some of the provided policies are irrelevant or noisy, and when the state observation is corrupted by sensing noise (i.e., partial observation).

Figure \ref{fig:baselines} illustrates the potential benefits of using a meta-policy in a fully observed driving example.
We construct a driving task where policies are trained on simulated cars with a steering bias (veers to one side) and a speed bias (goes faster than normal).
Given a novel task, which is a mix of these two regimes, we compare directly applying RL against different techniques that compose the two policies.
Results suggest that the meta-policy approach learns a composition that achieves a higher reward for the same amount of exploration compared to alternative composition techniques~\cite{wiering2008ensemble}.

\noindent \textbf{Our contributions are:}

\begin{enumerate} 
\item This paper explores the policy composition problem with Neural Networks in a driving simulator where the dynamics of the new task are an unknown composition of previously seen dynamics and the state observation is possibly corrupted by sensing noise.
\vspace{0.25em}
    
\item We evaluate the extent to which this meta policy learning approach efficiently learns a viable composition, is robust to ``distractor'' policies, and performs in degenerate cases, in comparison to alternative policy composition techniques and direct policy learning using RL.
\vspace{0.25em}

\item Our experimental results suggest that meta policy learning approaches are highly efficient in this new Neural Network setting since such policies can require large amounts of exploration.  
In a fully observed experiment, the meta policy learning approach achieves 2.6x of the reward by the next best policy composition technique and achieves this reward with 80\% less exploration than directly applying Reinforcement Learning (ignoring previously trained policies).
In a partially observed experiment, the meta policy learning approach converges after 50 iterations while the direct application of RL fails to converge even after 200 iterations.
\end{enumerate}

%% file: 2-relatedwork.tex
\section{Background and Related Work}
This section outlines the basic problem setting and related work.
We consider a finite-horizon Markov Decision Process (MDP), which is defined as a 5-tuple:
\[
\mathcal{M} = \langle S,A,T(\cdot,\cdot), R(\cdot,\cdot),N \rangle.
\]
$S$ is the state-space, $A$ is the action space, $T: S \times A \mapsto Prob(S)$ is the transition function that maps states and actions to a probability measure over subsequent states, $R: S \times A \mapsto \mathbb{R}$ is a reward function over the state and action space, and $N$ is the number of discrete time-steps (time-horizon). 
A policy $\pi: S \mapsto A$ is a function that maps states $S$ to actions $A$. 

A number of variants (and simplifications) of the policy composition problems have been studied in prior work.

\vspace{0.5em}
\noindent\textbf{Best-In-Set Problem: } Consider first the best-in-set problem where the agent must merely match the performance of the best policy in the set of given policies.
Suppose, we have a set of policies $\{\pi_1,...,\pi_k\}$ and an MDP $\mathcal{M}$, find the $\pi_i$ with the highest expected reward.
This problem can be solved with a Multi-Arm Bandit (MAB) algorithm, where each arm is a policy $\pi_i$ and each pull is rolling the policy out and evaluating the expected reward~\cite{jamieson2014lil}. 
Such ideas have been applied to robotics in works such as~\cite{matikainen2013multi,laskey2015multi, srivastava2014surveillance}, where MABs are used to select from a library of state-machines or motion plans.

We 
focus on the case where the dynamics of the new scenario are unknown but composed of a mixture of previously seen dynamics.
As seen in Figure \ref{fig:baselines}, none of the policies individually may achieve a high reward, but a composition of them might do so.
However, we do evaluate meta policy learning in the degenerate case where there exists a strong single basis policy (Section \ref{sec:dist}), and we also discuss how a hybrid MAB/RL formulation may apply in future work (\secref{disc}).

\vspace{0.5em}
\noindent\textbf{Ensembled Policies: }
Another approach to the composition problem is to have a \emph{passive} switching condition, i.e., one that is not learned from exploration.
For example, we can have a voting based system where all of the policies are queried at each state, and their actions are aggregated~\cite{wiering2008ensemble}, in particular, actions are stochastically drawn proportional to the number of constituent policies that selected that action.
In our simulated driving domain, we compare the meta policy approach to voting and find that the meta policy learning approach achieves 2.6x the reward.
Another approach is a confidence-based switching rule, similar to that in~\cite{chernova2007confidence} but applied in the context of LfD, where the robot selects the policy that has a higher confidence in that region of the state-space.
In our simulated driving domain, we find that the meta policy learning approach achieves 8.7x the reward to this approach.

\vspace{0.5em}
\noindent\textbf{Hierarchical Reinforcement Learning (HRL): }
 In HRL, policies can be constructed with closed loop sub-policies called options~\cite{kaelbling1993hierarchical, barto2003recent, comanici2010optimal, konidaris2009efficient, taylor2009transfer, dietterich2000hierarchical}.  Formally, an option is defined as a three-tuple of $\langle I,\pi, B \rangle$,
where $I$ is the initiation set - the set of states where the option can be taken; $\pi: S \to A$ is a specific policy taken under this option, and $B: s \to [0, 1]$ is a termination condition. An option can only be taken at $s_t$ if $s_t \in I$. 
The policy composition problem is a special case of HRL where options terminate after each time-step (also called one-step options).

The HRL approach for policy composition has been applied since the 1990s~\cite{kaelbling1993hierarchical, kalmar1998module, asada1994coordination, konidaris2009efficient, lin1993hierarchical, morimoto2001acquisition}.
We highlight a few of the representative works.
Kaelbing et al. studied voronoi decompositions of a continuous state-space to discretize an RL task~\cite{kaelbling1993hierarchical}.
Kalmar et al. and Asada et al.~\cite{kalmar1998module, asada1994coordination} explored the discrete-state, discrete-action case experimentally on robots.
Konidaris et al. studied composing segmented skills to perform a task \cite{konidaris2009efficient}. 

We take inspiration from this prior work and evaluate how these approaches can be incorporated into a Deep RL framework.
One of the new opportunities with Deep RL are Recurrent Neural Networks (RNN).
These networks efficiently store temporal state and can capture policies that depend on history.
This allows us to handle some problems with partial observation, e.g., sensing noise, which has always been a challenging problem in RL research~\cite{cassandra1998survey}.

\vspace{0.5em}
\noindent \textbf{Transfer in RL: }
Transfer Learning in RL has also been studied where the training happens on one task and then the goal is to apply the previous information to reduce training time in a novel scenario (see the survey by Taylor and Stone~\cite{taylor2009transfer}).
One of the first works that studied this problem was Selfridge et al.~\cite{selfridge1985training}, who explored control for the inverted pendulum by first training on a lighter pole.
The key difference is that Selfridge et al. studied the case where experience from prior training is transferred between tasks using a Q function and not policies, and others adopted similar transfer models~\cite{asada1994vision, singh1992transfer}.
We hope to explore such models in more detail in future work, but this study focuses on a transfer model where policies are given as input.

\subsection{Problem Statement}
This paper explores the following problem in continuous-state, discrete-action driving simulator:

\begin{problem}[Policy Composition Problem]
Given a set $\Pi = \{\pi_1,...,\pi_k\}$ \textbf{basis} policies defined over the same continuous state-space $S$ and discrete action space $A$. 
Let $\pi^{meta}$ be a meta policy, i.e., a policy that given a state $s \in S$ selects exactly one $\pi \in \Pi$:
\[\pi^{meta}: S \mapsto \{1,...,k\} \]
For an MDP $\mathcal{M}$ with unknown dynamics, but known to be a mixture of previously seen dynamics used to train $\Pi$, and a given reward function $R(s,a)$, find the meta policy that optimizes the reward. 
\end{problem}


%% file: 3-fo-case.tex
\section{Fully Observed Algorithm}\label{sec:fo}
First, we describe the technique used to optimize the meta policy when the MDP is fully observed.
This approach follows from prior work in Hierarchical RL~\cite{barto2003recent}, but we describe how we use a Multi-Layer Perceptron to represent the meta policy.

\begin{figure}[t]
\vspace{-10pt}
\centering
 \includegraphics[width=\columnwidth]{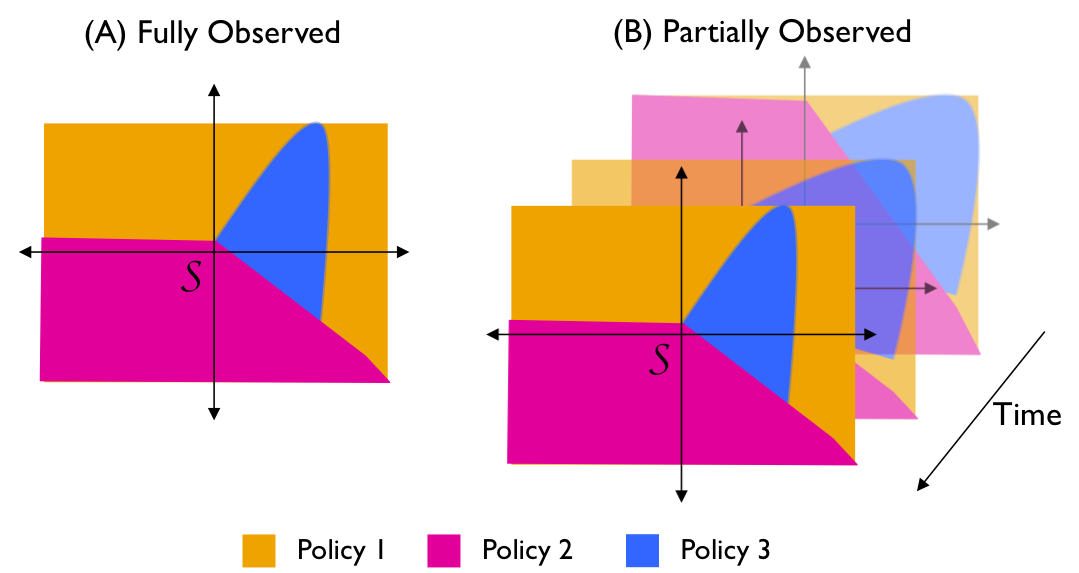}
 \caption{(A) In fully observed case, meta policy can be thought of as partitioning the state-space $S$ into distinct regions where each of the $k$ basis policies is applied. (B) In partially observed case, meta policy is partitioning the history of the task $S^t$ into such regions.}
 \label{fig:concept}
 \vspace{-15pt}
\end{figure}

\subsection{Parametrized Meta Policies}
In general, $\pi^{meta}$ can be difficult to represent exactly.
One option is to consider parametrized functions $\pi^{meta}_\theta$ where $\theta \in \Theta$ represents one function in a parameter space.
Let $\pi^{meta}_\theta(s)$ be a parametrized function mapping states to actions.
Then, when $\pi^{meta}_\theta(s)$ is applied to $M$ generates a distribution over $N$-step trajectories. Each $\pi^{meta}_\theta(s)$ has an expected reward:
\begin{equation} R(\theta) =  \mathbf{E}_{(s_n,a_n) \sim \pi^{meta}_\theta(s)}\left[\sum_{n=0}^N R(s,a)\right]
\label{eq:obj}
\end{equation}

However, the choice of parametrization is important.
A meta policy is a function that maps states to one of $k$ basis policies.
This can be thought of as partitioning the state-space into regimes associated with each policy (Figure  \ref{fig:concept}A).
However, these regimes may not define simply connected regions and actually may be quite complex.
There may be multiple regions of the state-space that correspond to the same basis policy.
In the cruise control example, it could be that a medium age car behaves like a new car at low speeds when driving straight (fewer effects of wear and tear), and high speeds when turning (turning dynamics dominate any small issues).

One solution is to use Neural Networks (NN) architectures to parametrically encode  $\pi^{meta}_\theta$ and then optimize $\theta$ with respect to the reward in Eq.~\eqref{eq:obj}. 
For the fully-observed case, we apply a Multi-Layer Perceptron Neural Network.
This network has two hidden layers (32 x 32), both with tanh nonlinearity functions and a softmax output layer to account for the domain's discrete action-space. 

\begin{figure*}
\vspace{-10pt}
\centering
 \includegraphics[width=0.9\linewidth]{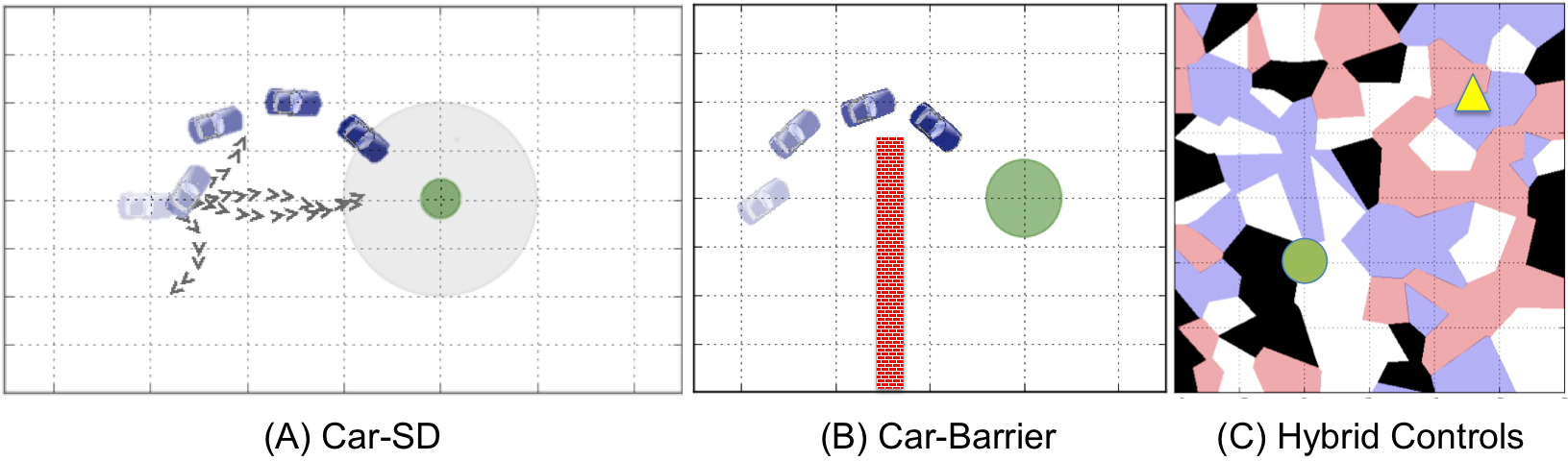}
 \caption{\textbf{Domains}: In both Car-SD and Car-Barrier, agent observes $(x, y, v, \theta)$ state tuple. 
 (A) In Car-SD, the car is in a 7 x 4 map and is tasked with driving to the goal state. The environment has two disjoint regions with different dynamics - one dynamics in a 1 unit radius (shaded) around the goal state, and the other dynamics in all other parts of the map. The dynamics of the domain are different in training basis policies than when training the meta policy. 
 (B) In Car-Barrier, the car is in a 5 x 4 map and needs to navigate around the barrier to reach the goal. 
 The dynamics are uniform dynamics throughout this map. (C) In the Hybrid Control environment, the agent is in a 12 x 12 map and needs to navigate from its starting position (triangle) to the green end goal through disjoint partitions of different dynamics. The partitions are formed from a voronoi mapping of 100 random points, each assigned with one of 4 dynamics as colorcoded above. }
 \label{fig:domains}
 \vspace{-15pt}
\end{figure*}

\subsection{Trust-Region Policy Optimization (TRPO)}
To learn the meta policy, we apply a version of a reinforcement learning algorithm called policy gradient.
Policy gradient algorithms optimize the objective in Equation \ref{eq:obj} with gradient ascent:
\[
\theta^{(i+1)} =\theta^{(i)}  + \lambda \cdot \nabla_{\theta^{(i)}}R(\theta^{(i)})
\] 
Since, we do not have access to the true distribution, only samples, we replace $R(\theta)$ with an empirical estimate $\tilde{R}(\theta)$:
\[
\theta^{(i+1)} =\theta^{(i)}  + \lambda \cdot \nabla_{\theta^{(i)}}\tilde{R}(\theta^{(i)})
\] 
Policy gradient is initialized with a parameter $\theta^{(0)}$ and ``rolls out'' $p$ trajectories from the policy $\pi^{meta}_{\theta^{(0)}}$ to calculate an estimated gradient of the reward function.
However, the gradient estimates can have high variance, so one solution is to optimize this efficiently is to use a technique called Trust Region Policy Optimization~\cite{schulman2015trust}.
Trust region methods take gradient steps in a small neighborhood around the current parameter value, clipping the size of the step if it exceeds a specified step-size parameter, preventing excessive oscillation due to noise.

%% file: 4-po-case.tex
\section{Partially Observed Algorithm}\label{sec:po}
Next, we describe the partially observed setting and the meta policy learning algorithm with RNNs.

\subsection{Partial Observation}
The main challenge with partial observation is that optimal policies depend on history, not just the current state. 
In RL, any amount of un-modeled observation noise (as opposed to process noise) is considered partial observation.
One of the new opportunities with Deep RL are Recurrent Neural Networks (RNN).
These networks efficiently store temporal state and can capture policies that depend on history.
We explore two scenarios: (1) meta policy is memory-less (MLP from previous section), and (2) meta policy is history dependent (RNN). Both scenarios use basis policies that are history-dependent (trained with an RNN).
Figure~\ref{fig:concept}b illustrates the difference between a history-dependent and memory-less meta policy.

\subsection{Overview of Recurrent Networks}
We briefly review the RNN architecture we use to represent both the basis policies and meta policies.
As cyclic neural networks, RNNs are able to capture temporal events that depend on the past.
They extract and encode relevant signal from the history in low dimensional embedding and are therefore particularly useful in the context of partial observations.

One challenge with RNNs is the \emph{vanishing gradient} problem, where dependencies over long time-scales cannot be modeled well.
The RNN architecture that we use is a Gated Recurrent Unit (GRU).
A gated recurrent unit (GRU) was proposed by~\cite{chung2014empirical} to capture dependencies of different time scales and address this issue. 
GRUs require less data to train than the popular (Long Short Term Memory) LSTM solutions to the same problem.
In all cases, the meta policy was trained using Trust Region Policy Optimization (step size of 0.001 and a linear feature baseline).

%% file: 5-exp-setup.tex
\section{Experimental Setup}\label{sec:res}

\subsection{Experimental Settings}

We implement meta policy learning with RLLab~\cite{duan2016benchmarking}, which is a library for Deep RL. The library provides a comprehensive set of policy optimization algorithm implementations along with a suite of continuous control tasks for evaluation. We use the RLLab implementation of Trust Region Policy Optimization.  

\noindent \textit{Evaluation Metrics} --
We use the average reward as the number of iterations increase as a metric of evaluation and comparison of meta policy against other baselines. Another metric we used to evaluate the framework was the number of iterations needed until the mean reward exceeds a given reward threshold with respect to the number of basis policies. 

\subsection{Experimental Environments}

In two of our evaluation platforms, we used a remote control car simulator \cite{geramifard} in which a car learns to drive to a pre-defined target. 
We chose this domain because of its challenging non-holonomic movements, ease of visualization, intuitive behavior, and 
its ability to relate to
highly complex problems with continuous state spaces.
The remote control car domain has a four-dimensional state-space of $x, y, v, \theta$, where $(x, y)$ are the position, $v$ is the velocity, and $\theta$ is the direction of the car. 
For any state, the car can take one of 9 actions, formed from pairs of (+0.03, 0, -0.03) velocity and (+30, 0, -30) degree turning angle. 
The main intention is to evaluate and characterize the reinforcement learning agent that chooses amongst policies trained on different dynamics.

\noindent \textit{1. Car with Split Dynamics (Car-SD) } --
In this environment (\figref{domains}a), the agent is in a 7 x 4 map and is tasked with driving to the goal state. We use two different dynamics models, each located in a disjoint partition of the state space. The car is placed three units away from the goal which has a success radius of 0.2 units with a constraint that the car must be going slower near the goal - even in the region of the goal, it must have an absolute velocity less than 0.2 units in order to succeed.

One dynamics model $D_1$ has a constant position bias (modeling slippage) and the other model $D_2$ with a constant turning angle bias (modeling alignment issues). The partition with $D_1$ is located in a 1 unit radius around the goal position, and the rest of the environment has $D_2$. 
In this environment, the car needs to drive to the goal while passing through regimes with different dynamics. 
The time horizon is set as 500 steps. 
The car receives a penalty proportional to the distance from the goal for each step taken.
The car receives a reward of -10 when it collides with the environment boundaries (walls) and a reward of 100 for reaching the goal.
The meta policy for this environment uses two basis policies that are each trained on duplicates of this environment layout but with only one dynamics model present.

\noindent \textit{2. Car with Barrier (Car-Barrier) } --
We also construct a variant of the Car domain with obstacles. 
In this environment (\figref{domains}b), the car is placed in a 5 x 4 map, 3 units away from the goal state which has a success radius of $0.4$. 
The obstacle environment ensures that there is a high penalty for taking a random action in an inopportune place, like right next to a barrier.
This allows us to better characterize the policy recovery problem, where one policy can navigate around the obstacles and the others are random noise.
We use a uniform dynamics model for the entire environment. 
The time horizon for the task is set as 200 steps. 
The car receives a step penalty of -1 for each step taken; a reward of 500 when inside the radius of the goal and a penalty of -100 when colliding with another structure (i.e., the barrier).

\noindent \textit{3. Hybrid Control Environment} --
In this environment (\figref{domains}c), the agent is in a 12 x 12 map and needs to navigate from its starting position to the goal state. We generate $N$ linear dynamics models (A, B for $x_{t+1} = Ax_{t} + Bu_{t}$), where $N$ is the number of controllers we will be giving the meta policy. For each (A,B), we solve for a gain matrix K that will be used as the controller ($x_{t+1} = (A - BK)x_{t}$). 
We take the 4 of the $N$ dynamics models and use them exclusively in the different dynamics partitions. The partitions are formed from a voronoi mapping of 100 random points, each assigned with one of 4 dynamics. 
The reward function is a negative quadratic reward, and Gaussian noise is added.
The agent begins at an initial state and needs to reach within a 0.4 radius around the goal state within 100 time-steps.



%% file: 6-exp-results.tex
\section{Experimental Results}
We present results in two domains, a simulated car driving domain (discrete actions) and hybrid linear control domain (continuous actions).

\begin{figure}[t!]
\centering
\vspace{-5pt}
 \includegraphics[width=0.85\linewidth]{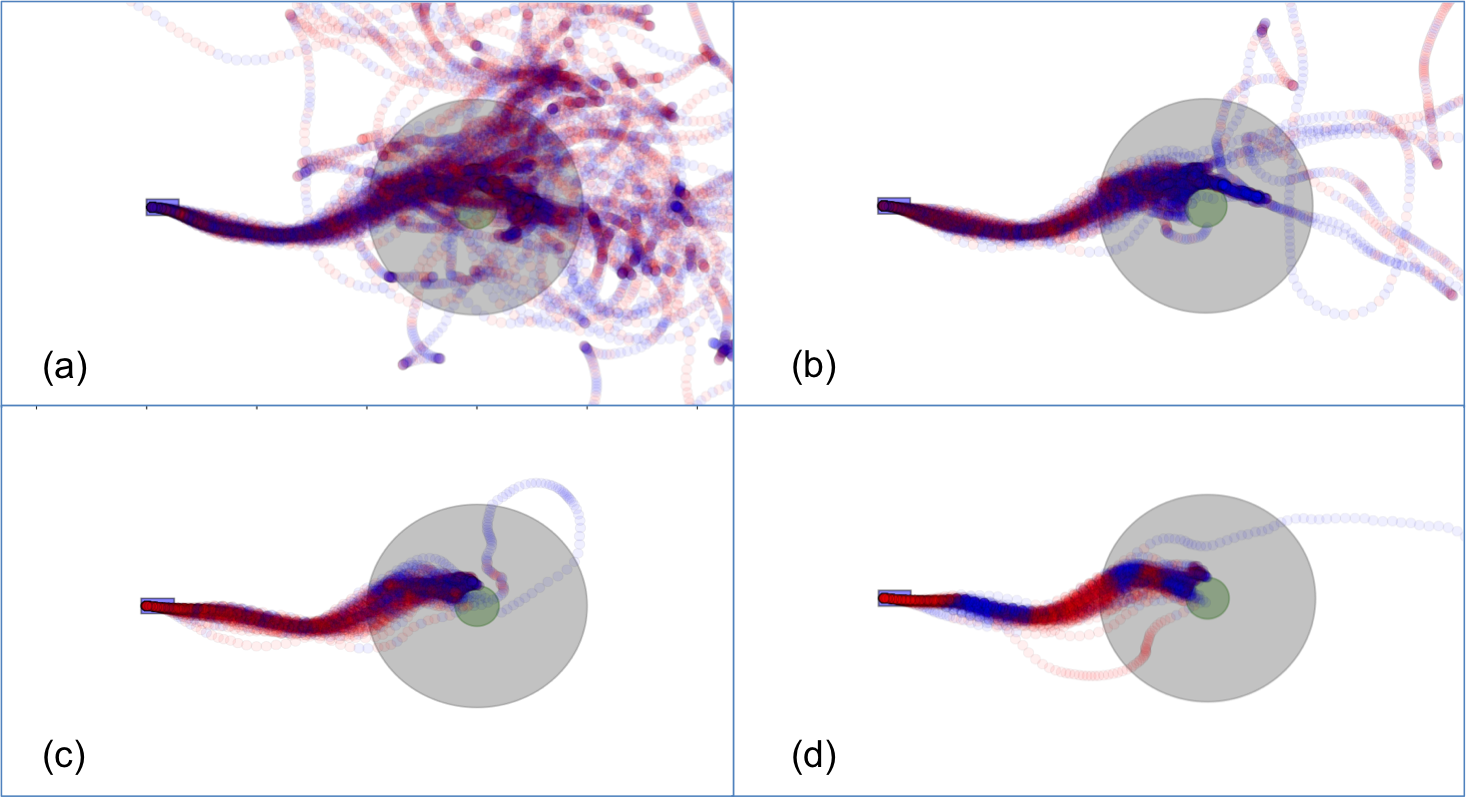}
 \caption{Car-SD: Above are the traces of 100 rollouts, visualizing the mapping from state to subpolicy. From (a) to (d), each map shows the current policy at 10, 50, 100, 500 iterations of training respectively. A policy with clear segments is drawn as the iterations increase, but it is interesting to note that the multiple policies are used in one dynamics region, including those that were not originally trained on that dynamics model.}
 \label{fig:stitched}
 \vspace{-15pt}
\end{figure}

\begin{figure*}[ht]
\centering
 \includegraphics[width=0.5\linewidth]{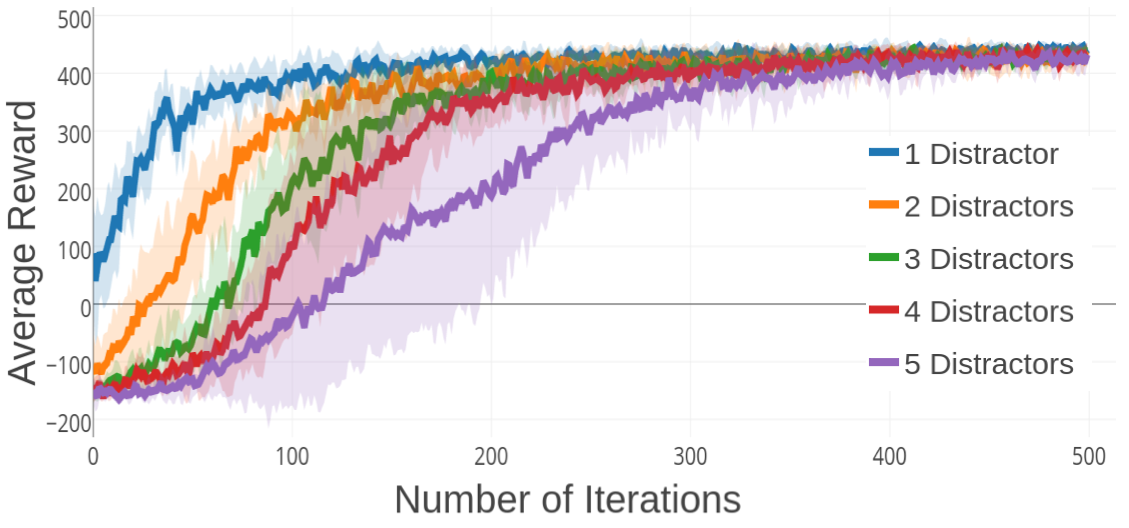}
\quad
  \includegraphics[width=0.3\linewidth]{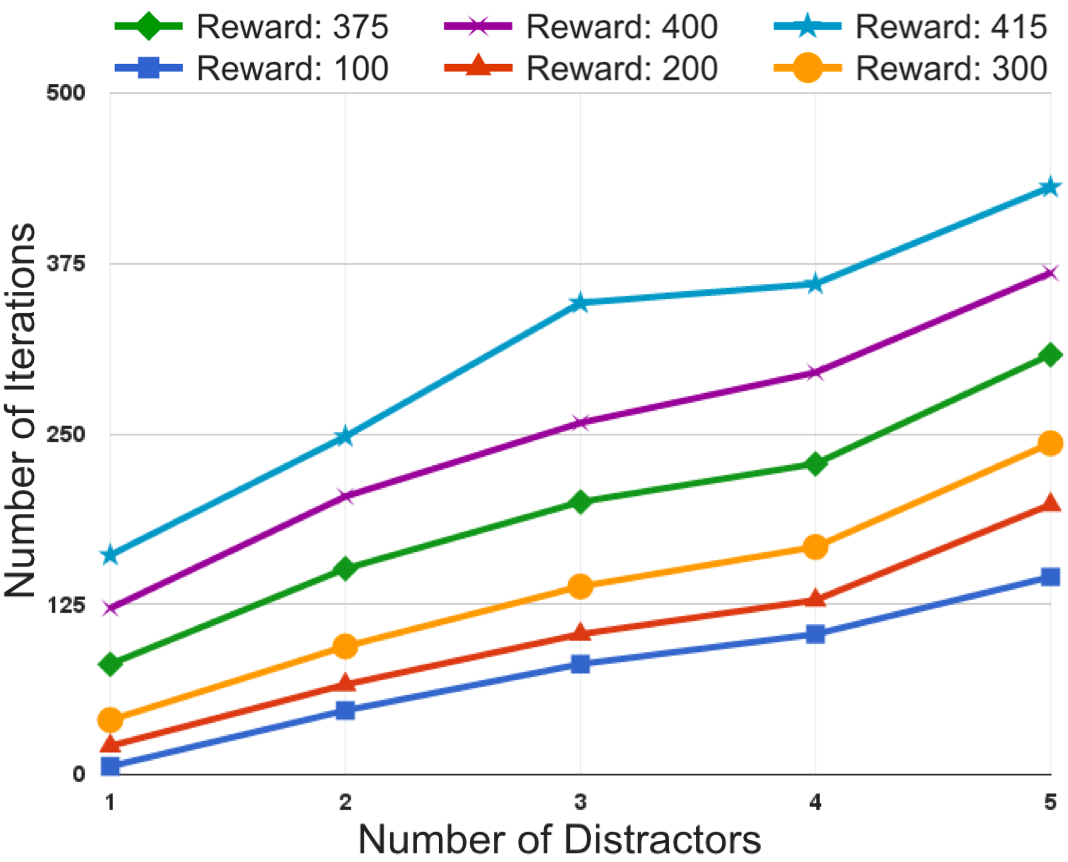}
 \caption{\textbf{Distractors in Car-Barrier}: (a) We evaluate meta policy learning in its ability to reject distractor (random) policies. We increase the number of distractor policies, while maintaining one optimal policy and measure the number of iterations to convergence to several different reward levels. 
 (b) Results suggest that it is possible to consistently recover a successful policy, even as the number of noisy agents are varied for roughly linearly more iterations.}
 \label{fig:recovery_chart}
\end{figure*}

\subsection{Baseline Comparison}
In the first experiment, we test the following hypotheses: (1) meta policy learning learns a composition that achieves a higher reward than alternatives, (2) training a new policy with RL, disregarding the previously trained policies, requires more exploration to achieve the same reward as meta policy learning.
We evaluate the frameworks on the Car-SD domain, which has two dynamics stitched together - $D_1$ (position bias) and $D_2$ (angle bias).
Two basis policies for the meta policy are trained on 7 x 4 uniform dynamics maps with the start state and goal state at the same position. One policy is trained on a map with only $D_1$ and the other with only $D_2$. Training is done with Q-Learning ($\alpha=0.95, \lambda = 0.4$) with a randomized Radial Basis Function Q-Function approximator of 2000 randomly placed kernels. 


\noindent We evaluate meta policy learning against the following baselines:

\begin{figure}[t!]
\vspace{-10pt}
\centering
 \includegraphics[width=0.8\linewidth]{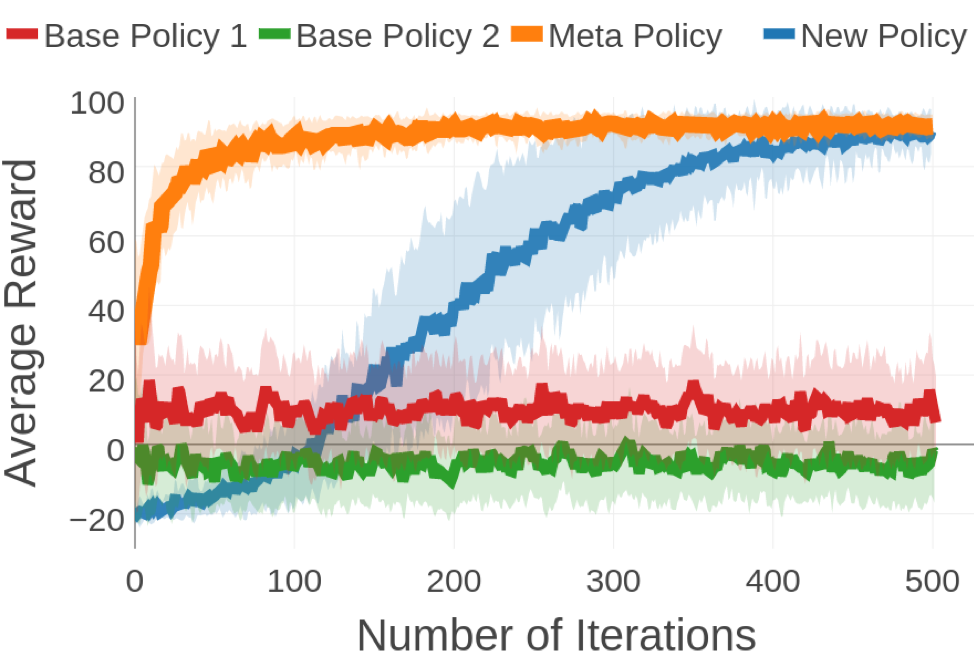}
 \caption{\textbf{Distinct Dynamics}: Given only basis policies, the agent will not successfully reach the goal. Using TRPO, Meta RL quickly converges to a successful composition of the basis policies, significantly faster than training a policy from scratch.}
 \label{fig:distinct_dynamics}
 \vspace{-15pt}
\end{figure} 

\begin{enumerate}
\item\textbf{Basis Policies: } We evaluate the reward earned by each of the individual basis policies if no more learning occurs.
\item \textbf{Voting: } We apply an ensembling technique where actions are stochastically drawn, proportional to the number of constituent policies that selected that action~\cite{wiering2008ensemble}. 
\item \textbf{Transition Estimator: } We motivate this example by using state transition history as a heuristic for applying the optimal basis policy for a certain region. Given that the policies are trained on dynamics, we generate 100 rollouts $T = [s_0, a_0, s_1, a_1, ... a_{n-1}, s_n]$. and fit a Kernel Density Estimator over $(s_i, a_{i}, s_{i + 1})$ transition tuples. On the stitched domain, each agent is queried for a confidence estimate, and the action is stochastically drawn, proportional to the confidence estimate.
\item \textbf{New Policy: } We apply RL directly to the new domain and evaluate the number of iterations required before achieving a given reward.

\end{enumerate}
Figure \ref{fig:baselines} illustrates the results.
First, each of the basis policies applied individually does not achieve a high reward on the mixed domain.
The ensembling techniques, voting and confidence, perform marginally better than the basis policies.
Since these techniques are not adaptive, they have a fixed level of performance.
The voting based estimator achieves 3x the reward of the best basis policy while the transition estimator achieves 10\% better than the best basis policy.
Next, we evaluate the adaptive approaches, meta policy learning and directly applying reinforcement learning.
Meta policy learning converges after 100 iterations of TRPO (with a batch size of 1000, discount factor of 0.995, and step size of 0.001).
In comparison, directly learning a policy requires 5x times as many iterations to converge to the same reward.
In this case, having access to previously trained policies on the independent dynamical regimes avoids nearly 8000 trajectory rollouts.

Figure \ref{fig:distinct_dynamics} illustrates the meta policy's convergence.
Initially, the meta policy explores by trying both basis policies in different states (Figure \ref{fig:stitched}a).
As the meta policy converges, it learns when to apply each policy (Figure \ref{fig:stitched}d).
Figure \ref{fig:stitched}d is interesting as it shows that even though there are only two dynamical regimes in the domain, the typical trajectory has three switches.
In RL problems, due to the delayed reward, it is sometimes beneficial to take a locally suboptimal action (i.e., applying a policy trained on a different dynamics regime) to achieve a higher ultimate reward.
In this case, the car leverages the angle-biased policy to guide itself to the right orientation.

\begin{figure*}[t]
\vspace{-10pt}
\centering
 \includegraphics[width=0.38\linewidth]{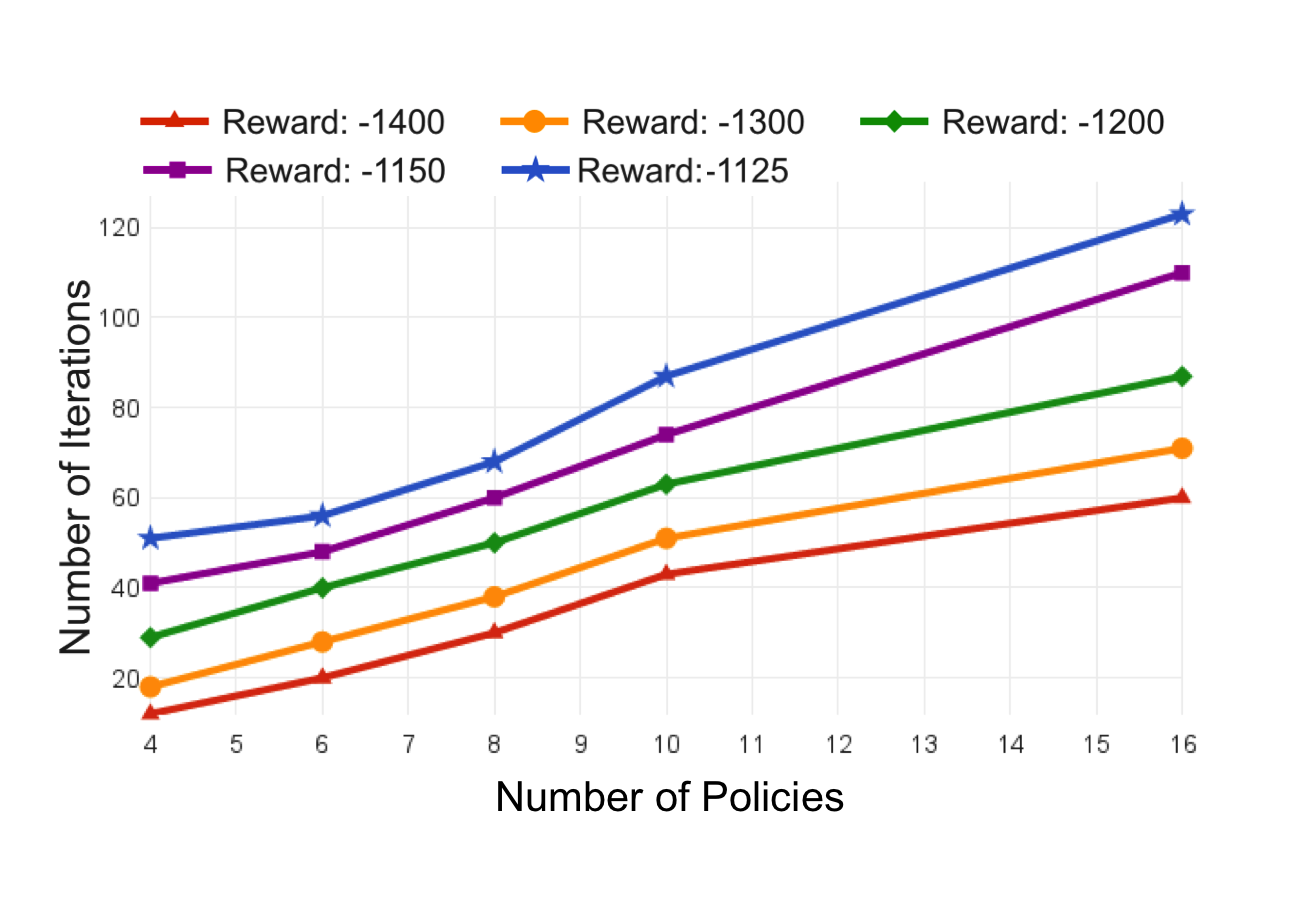}
 \quad
 \includegraphics[width=0.38\linewidth]{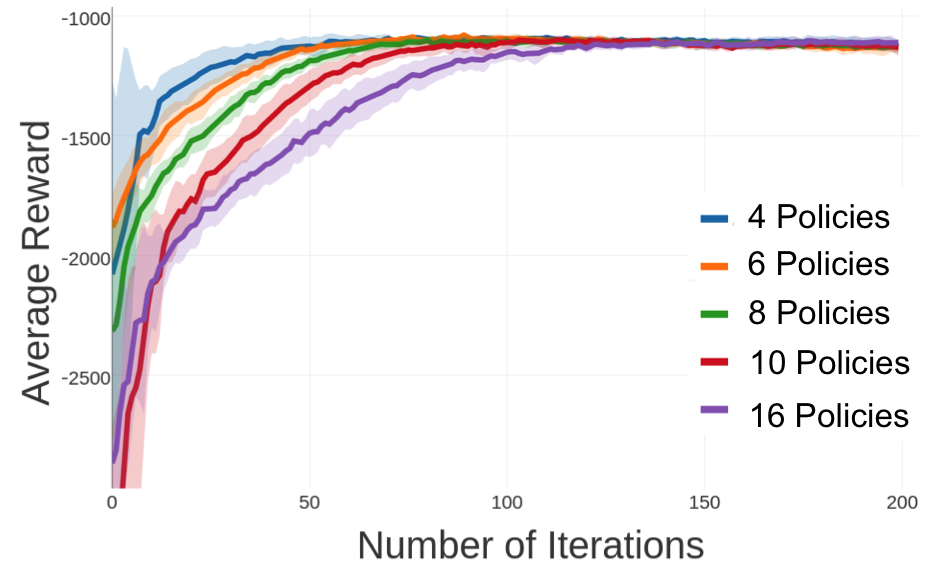}
 \caption{\textbf{Multiple policies in Hybrid Control}: Meta Policy Learning not only converges in the hybrid control example but can also reject extraneous policies as in the discrete-action case.}
 \label{fig:ctrl_convergence}
 \vspace{-15pt}
\end{figure*}

\begin{figure}[ht]
\centering
\includegraphics[width=\linewidth]{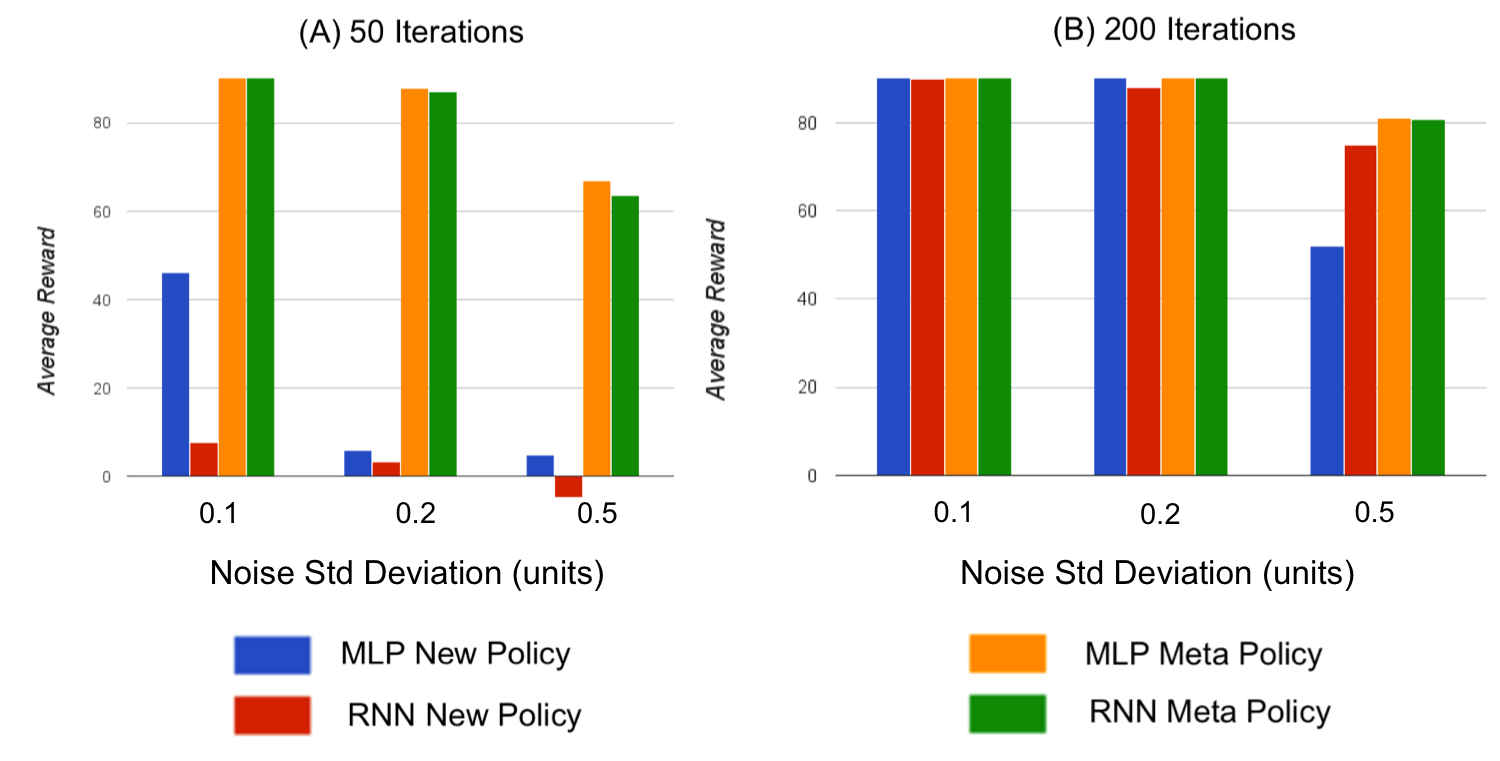}
\vspace{-10pt}
\caption{\textbf{Sensitivity to Noise} We evaluate the Car-SD domain with sensing noise and vary the magnitude of noise. The meta policy learning approach converges in 50 iterations, while after 200 iterations the new policy approach has not yet converged. Surprisingly, we find that a memory-less meta policy (using a MLP) instead of an RNN has a faster convergence rate. \label{noisyobservations}}
\vspace{-15pt}
\end{figure} 

\subsection{Sensitivity to Noise}
Next, we evaluate the Car-SD domain with sensing noise, a form of partial observation.
We evaluate the convergence of meta policy learning as a function of the probability and magnitude of noise.
We evaluate: (1) training a history-dependent meta policy (RNN), (2) training a memory-less meta policy (MLP), (3) training a new history-dependent meta policy with RL (RNN)  (4) training a new memory-less meta policy with RL (MLP). We use TRPO for all training, and we train all policies with a batch size of 4000, discount factor of 0.995, and step size of 0.001. 

Figure \ref{noisyobservations} displays the effect of noise on policy performance on the new policies.
This illustrates that history dependent policies achieve a higher reward for larger amounts of noise.
History-dependent policies are not strictly better since they inherently have a larger number of parameters and require more data to converge.
For small amounts of noise, this tradeoff is in favor of the memory-less policies.
On the other hand, as the noise rate increases, history-dependent policies are more accurate.

The surprising result is when we consider the meta policies.
As in the fully observed case, meta policy learning converges to the maximum possible reward with less exploration than RL--exploiting the previously trained policies.
The benefits of meta policy learning are even more pronounced under large amounts of observation noise.
We find that even with 4x as many iterations, directly applying RL to learn a new policy (memoryless or history-dependent) fails to achieve the same performance as meta policy learning.

\subsection{Distractor Policies}\label{sec:dist}
We now evaluate the degenerate case of meta policy learning, where out of the set of $k$ policies, there is one viable policy, and all of the others are noise (take random actions)--and we call these random policies \emph{distractors}.
We evaluate two hypotheses: (1) meta policy learning will recover the near-optimal policy, (2) the amount of exploration to recover the policy is roughly linear in the number of distractors. 
These experiments were run on a variant of the car domain, Car-Barrier.

The viable basis policy was learned using Q-Learning over 1000 episodes with a Radial Basis Function approximator.
We plot the results in Figure \ref{fig:recovery_chart}a-b.
Results suggest that the algorithm can recover the viable policy despite with multiple distractors. 
With up-to 5 distractors, the meta policy trained with TRPO converges to the reward of the viable policy with no more than 500 iterations (with a batch size of 2000, discount factor of 0.995, and step size of 0.001).
We also see that increasing the number of agents delay the convergence rate in a roughly linear fashion. 
Figure \ref{fig:recovery_chart}a shows the number of iterations to achieve a fixed reward.

\subsection{Hybrid Control}
Next, we evaluate meta policy learning in a continuous action space using the Hybrid Control environment. 
In spirit, this experiment is similar to the distractor policy experiment, but instead of random controllers, we use sub-optimal controllers.
The experiment setup has disjoint partitions of 4 different dynamics models. No single controller can bring successfully reach the goal by itself.  We run meta policy learning (with a batch size of 1000, discount factor of 0.995, and step size of 0.001) using up to 16 different controllers as the set of basis policies, where 4 of the controllers always being solutions to the four environment dynamics models. From \figref{ctrl_convergence}, we also observe a linear order of convergence.

%% file: 7-conclusion.tex
\section{Conclusion}
This paper presents an initial exploration of how deep reinforcement learning can be used to address the policy composition problem.
We present results in a simulated driving domain consisting of varying dynamics mixes, distractor policies, magnitudes/distributions of sensing noise, and obstacles.
In the fully observed setting, the meta policy learning approach achieves 2.6x of the reward by the next best policy composition technique and achieves this reward with 80\% less exploration than directly applying Reinforcement Learning (ignoring previously trained policies).
In the partially observed setting, meta policy learning converges to high-reward policies in roughly 50 iterations while the direct application of RL fails to converge even after 200 iterations.

%% file: 6a-discussion.tex
\section{Discussion \& Future Work}
\seclabel{disc}

\begin{figure}[t]
\centering
\includegraphics[width=0.9\columnwidth]{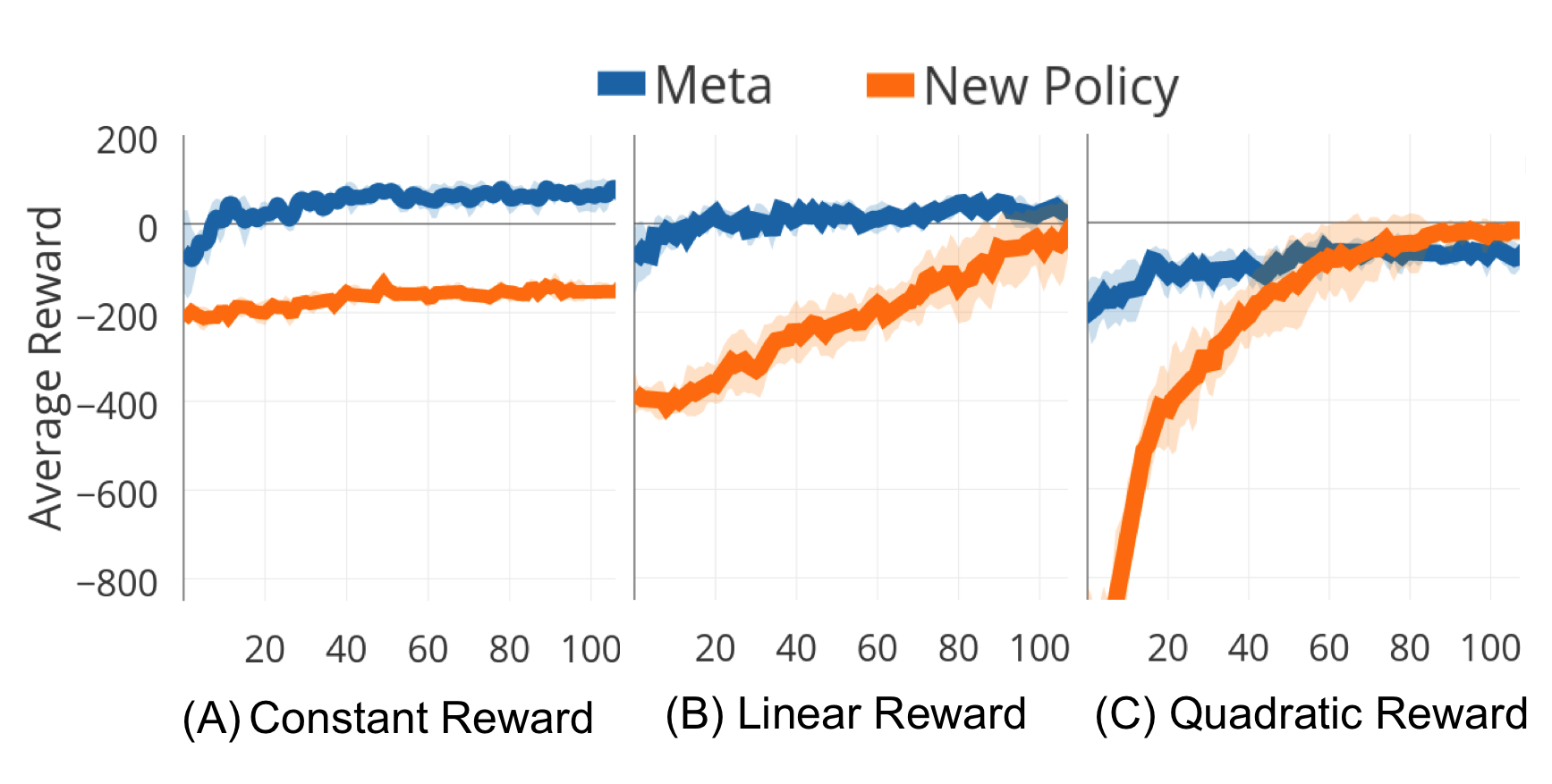}
\caption{Policy convergence graphs with respect to different reward strengths on Car-SD. We vary the strength of the reward in each experiment, (A) using a constant reward of -1, (B) a negative linear reward proportional to euclidean distance to goal, and (C) a negative quadratic reward.}
\label{fig:recovery_good}
\vspace{-15pt}
\end{figure} 

It is well-known that the convergence rate of RL is highly dependent on how rewards are shaped~\cite{ng1999policy}. 
A well-shaped reward can serve to guide the learner towards the optimal policy without much exploration.
However, it may not be possible to design a well-shaped reward with the available data especially in sequential tasks where success depends on satisfying a sequence of conditions~\cite{krishnan2016hirl}.
We provide a rough characterization of the effects of rewards in the Car-SD domain, shown in \figref{recovery_good}. There is a clear increase in convergence rate of training the new RL policy as we increase the strength of the reward. Conversely, the evidence suggests that meta policy learning would provide the greatest benefit when rewards are sparse or delayed.
Exploring the link between reward shaping and hierarchical reinforcement learning is an important avenue of future work.

One of the key arguments against meta policy learning is the potential optimality gap. Meta policy learning's objective is to find the best composite policy, which can achieve a significantly lower reward than the optimal policy for the MDP.
Another direction for future work is to study this gap in further detail.
We will explore hierarchical MDP approaches that allow for learning in the basis policies, perhaps reducing this potential gap.
Another important theoretical direction is to quantify this gap analytically for common problem types.

In terms of the lower bound, meta policy learning also can fall short.
Consider the degenerate case when there exists a single policy that dominates throughout the state-space.
The best-in-set problem can be solved with a multi-arm bandits (MAB) approach.
When applied to the experiment in Section \ref{sec:dist} found that MAB approach is far more efficient than using RL.
We evaluated the distractor policy experiment using an Upper Confidence Bound strategy and found that with 3 distractor policies this approach applies the correct policy within 4000 steps - which is two orders of magnitude faster than using hierarchical RL. 
We hope to explore using MAB to initialize policy gradient iterations such that the meta policy learning iterations at least begin with a performance no worse than the best basis policy.